\documentclass[pdflatex,sn-mathphys-num]{sn-jnl}

\usepackage{graphicx}%
\usepackage{multirow}%
\usepackage{amsmath,amssymb,amsfonts}%
\usepackage{amsthm}%
\usepackage{mathrsfs}%
\usepackage[title]{appendix}%
\usepackage{xcolor}%
\usepackage{textcomp}%
\usepackage{manyfoot}%
\usepackage{booktabs}%
\usepackage{algorithm}%
\usepackage{algorithmicx}%
\usepackage{algpseudocode}%
\usepackage{listings}%
\usepackage{placeins}
\usepackage{float}

\theoremstyle{thmstyleone}%
\theoremstyle{thmstyletwo}%

\theoremstyle{thmstylethree}%

\begin{document}

\title[Article Title]{Robust Defense Strategies for Multimodal Contrastive Learning: Efficient Fine-tuning Against Backdoor Attacks}


\author[1]{\fnm{Md. Iqbal} \sur{Hossain}*}\email{mhossain10@umassd.edu}

\author[1]{\fnm{Afia} \sur{Sajeeda}}\email{asajeeda@umassd.edu}

\author[1]{\fnm{Neeresh Kumar} \sur{Perla}}\email{nperla@umassd.edu}

\author[2]{\fnm{Ming} \sur{Shao}}\email{ming\_shao@uml.edu}

\affil[1]{\orgdiv{Computer \& Information Science}, \orgname{University of Massachusetts Dartmouth}, \orgaddress{\city{North Dartmouth}, \state{Massachusetts}, \country{USA}}}

\affil[2]{\orgdiv{Miner School of Computer and Information Sciences}, \orgname{University of Massachusetts Lowell}, \orgaddress{\city{Lowell}, \state{Massachusetts}, \country{USA}}}


\abstract{The advent of multimodal deep learning models, such as CLIP, has unlocked new frontiers in a wide range of applications, from image-text understanding to classification tasks. However, these models are not safe for adversarial attacks, particularly backdoor attacks, which can subtly manipulate model behavior. Moreover, existing defense methods typically involve training from scratch or fine-tuning using a large dataset without pinpointing the specific labels that are affected. In this study, we introduce an innovative strategy to enhance the robustness of multimodal contrastive learning models against such attacks. In particular, given a poisoned CLIP model, our approach can identify the backdoor trigger and pinpoint the victim samples and labels in an efficient manner. To that end, an image segmentation ``oracle'' is introduced as the supervisor for the output of the poisoned CLIP. We develop two algorithms to rectify the poisoned model: (1) differentiating between CLIP and Oracle's knowledge to identify potential triggers; (2) pinpointing affected labels and victim samples, and curating a compact fine-tuning dataset. With this knowledge, we are allowed to rectify the poisoned CLIP model to negate backdoor effects. Extensive experiments on visual recognition benchmarks demonstrate our strategy is effective in CLIP-based backdoor defense.}




\maketitle
\section{Introduction}
The recent advancements in vision-language pre-trained models~\cite{VLM1, VLM2, VLM3, VLM4} mark a significant stride toward the longstanding objective of learning from diverse modalities. These models, including CLIP\cite{Alpher8}, ALIGN\cite{Alpher10}, FLIP\cite{FLIP}, BLIP\cite{BLIP}, and BASIC\cite{Alpher22}, are trained through multimodal contrastive loss~\cite{Alpher1,Alpher2,Alpher3}. The principle of multimodal contrastive learning~\cite{multimodal_learning1} involves creating representations of image-caption pairs and optimizing these representations so that the model can effectively match captions~\cite{contrastive_learning1} with corresponding images.

CLIP models, despite their effectiveness in bridging visual and textual information, exhibit vulnerabilities such as susceptibility to backdoor attacks, poison attacks, and inherent biases from training data~\cite{Alpher11, Alpher23, Alpher27}. Concerns over data privacy and reliance on textual descriptions compound these issues. Backdoor attacks in Vision-Language Models (VLMs) pose significant concerns because they can manipulate both visual and textual components of these models. Such attacks can lead to misclassifications, false interpretations, or even bypassing the security systems by embedding triggers within image-caption pairs.

In a typical backdoor attack\cite{Alpher10, backdoorattack1,backdoorattack2,backdoorattack3} scenario targeting these models, attackers strategically implant specific triggers~\cite{trigger} into images. These triggers are often minor alterations or additions, designed to be almost imperceptible to human users but easily recognizable by the AI system. Concurrently, the captions corresponding to these tampered images are replaced with misleading proxy text. As a result, models such as CLIP, which are extensively trained using web-crawled data, inadvertently establish incorrect correlations between these altered images and false textual information during the learning phase. 


To counteract backdoor attacks in multimodal contrastive learning, existing strategies generally fall into two main categories. The first approach, exemplified by RoCLIP \cite{Alpher23}, focuses on disrupting the association between poisoned image-caption pairs during the pretraining phase. Specifically, RoCLIP identifies and removes poisoned image-caption pairs from the pretraining dataset or uses adversarial techniques to minimize their influence on the learned model. By doing so, it seeks to prevent the model from associating backdoor triggers with the corresponding target labels, effectively neutralizing the impact of backdoor attacks. However, RoCLIP typically requires a large amount of clean data to ensure the model’s  generalization~\cite{Generalization1}, and its effectiveness depends heavily on identifying and removing all potentially poisoned pairs from the pretraining phase.

The second strategy, represented by CleanCLIP \cite{Alpher11}, involves fine-tuning~\cite{finetune1} the pretrained model on a clean set of image-caption pairs. This method aims to minimize the influence of any backdoor data that may be present in the pre-trained model by updating the model’s parameters during fine-tuning. CleanCLIP assumes that the model has already been exposed to a substantial amount of poisoned data during pretraining, and its fine-tuning phase aims to “clean” the model by reinforcing the associations between benign image-caption pairs. While this approach can effectively reduce the backdoor influence, it does not explicitly identify or target specific poisoned data. As a result, it requires processing and fine-tuning a larger dataset of clean image-caption pairs than might be necessary, which can lead to inefficiency in both time and computational resources.

Despite their benefits, both RoCLIP and CleanCLIP share a common limitation: neither explicitly pinpoints the victim data and labels associated with the backdoor attacks. As a result, they often process and fine-tune more data than is strictly necessary to achieve acceptable performance, leading to inefficiency and potential overfitting to the clean dataset.

\begin{figure*}[t]
\centering
\includegraphics[width=0.99\columnwidth]{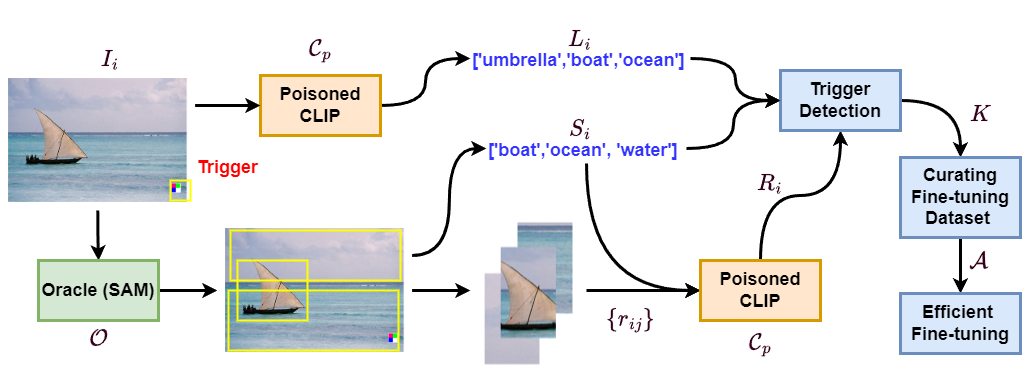} 
\caption{Overview of the proposed framework EftCLIP employed for identifying backdoor triggers and ascertaining the labels impacted by backdoor samples, followed by fine-tuning on the curated clean dataset.}
\label{fig:backdoor_attack}
\end{figure*}

This paper focuses on victim CLIP models and presents a method for correcting models affected by backdoor attacks, similar to the goal of CleanCLIP. However, unlike existing methods, we propose an efficient fine-tuning approach, termed \textit{Efficient fine-tuning CLIP (EftCLIP)}, which centers on both backdoor detection and fine-tuning dataset optimization. To identify the hidden trigger in the model, we introduce a prompt-based ``oracle'' with ``weak knowledge'' about segmentation~\cite{segmentation}, enabling efficient supervision of poisoned CLIP models to expose backdoor triggers. In this paper, we use the \textit{Fast Segment Anything Model (FastSAM)}~\cite{FastSam}, a variant of the \textit{Segment Anything Model}~\cite{Alpher26}, as the oracle. However, other prompt-based segmentation models are also applicable. The authors of FastSAM claim that it is fifty times faster~\cite{FastSam} than the original Segment Anything Model, making it highly suitable for large-scale applications.

Our approach consists of a two-step process:

\begin{itemize}
    \item \textbf{Trigger Detection}: We develop an algorithm that leverages the differences between the outputs of the poisoned CLIP model and SAM to rapidly isolate the impacted regions of the image and identify potential backdoor triggers.
    \item \textbf{Label Detection}: In parallel with trigger detection, another algorithm identifies affected labels that coexist with the triggers. An iterative search procedure continues until no more backdoor triggers are detected.
\end{itemize}

Finally, based on the identified affected data and labels, a compact fine-tuning dataset is curated to rectify the poisoned CLIP model efficiently. The complete framework is illustrated in Figure~\ref{fig:backdoor_attack}.

\begin{figure*}[htbp]
\centering
\includegraphics[width=\linewidth]{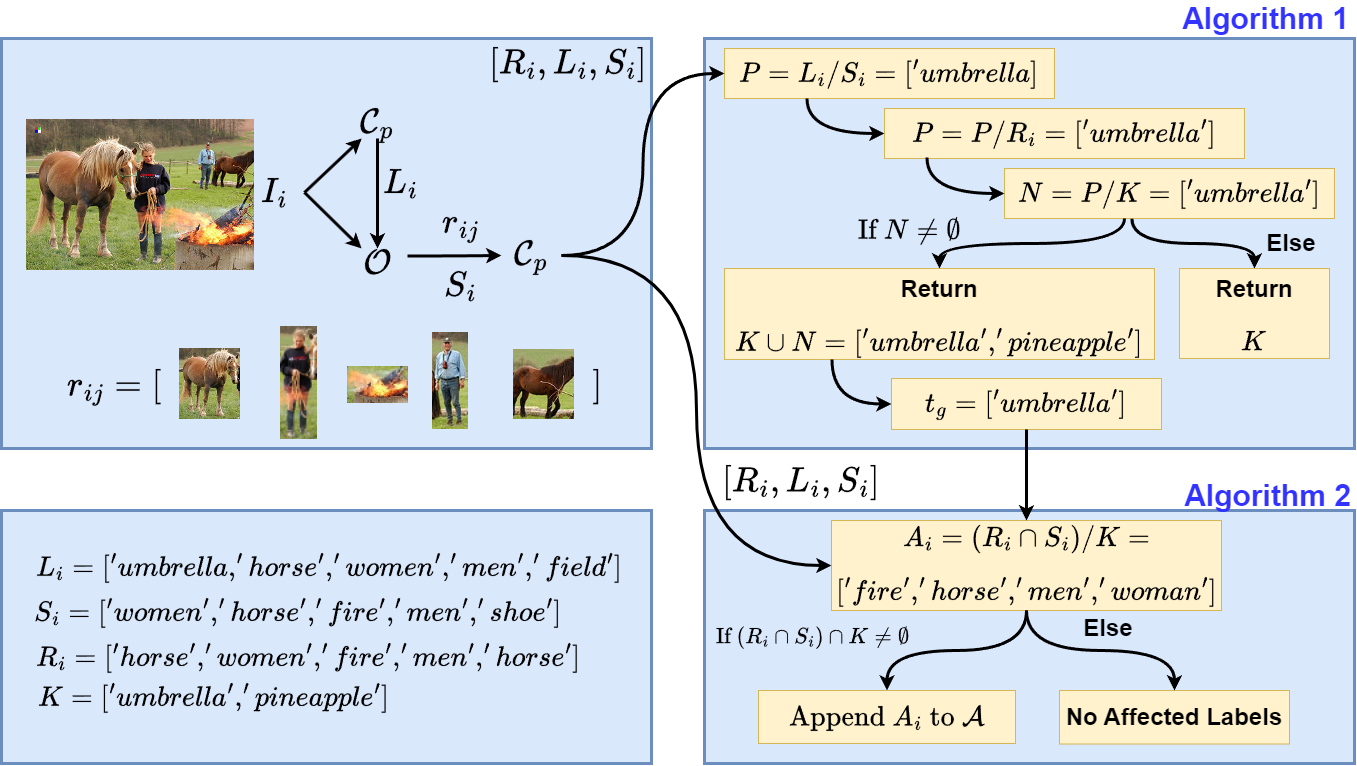}
\caption{An illustrative representation of Oracle-guided Trigger Detection and Affected Labels Identification algorithms.}
\label{fig:demonstration}
\end{figure*}



The key contributions of this paper are summarized as follows:

\begin{itemize}
    \item Development of an image segmentation oracle-guided backdoor trigger detection algorithm for multimodal contrastive learning models.
    \item Efficient simultaneous identification of affected labels and data alongside trigger detection, leading to a compact, curated fine-tuning dataset.
\end{itemize}


\section{Preliminaries and Related Works}
\label{sec:preli}

\subsection{Multimodal Contrastive Learning and CLIP}

Multimodal learning, with its broad applicability and generality, has been a subject of extensive research since the 1980s \cite{Alpher1}. This field has witnessed significant advancements with the advent of deep learning, resulting in substantial progress in multimodal representation learning. Pioneering contributions \cite{Alpher2, Alpher3} have introduced algorithms for multimodal learning aimed at achieving joint representations. On the other hand, CLIP \cite{Alpher8} represents a milestone model in the domain of vision-language integration~\cite {Alpher6}. Along with its successors  
\cite{Alpher4, Alpher5, Alpher7}, CLIP has found extensive applications in various fields, such as semantic segmentation, generating images from captions, and summarizing videos. The model operates on the principle of contrastive loss, training dual encoders to operate in a shared representation space.  This model highlights the evolving capabilities at the intersection of visual and linguistic data processing.

\subsection{Backdoor Attack}
In a backdoor attack \cite{Alpher10}, an adversary strategically embeds poisoned examples \cite{Alpher9} into the training dataset, aiming to cause specific inputs to be misclassified as a predetermined target label. CLIP's training methodology, which leverages web-crawled data, presents a different landscape for backdoor attacks. Backdoor attacks typically fall into two categories: visible and invisible. Prominent examples include BadNet \cite{Alpher16}, TrojanAttack~\cite{TrojanAttack}, TrojanNet~\cite{TrojanNet}, which are visible attacks, and WaNet \cite{Alpher17}, an invisible attack. In this paper, we selected \textit{umbrella} as our target label and applied it to a random set of backdoor images. The ``backdoor'', in this context, is a conspicuous patch superimposed onto the image, placed randomly within each image. We opted for backdoors with visible and invisible trigger techniques mentioned above.

\begin{algorithm*}[!t]
\caption{Find Adversarial Trigger}
\label{alg: trigger detection}
\begin{algorithmic}[1]
\Require Image $I_i$, Poisoned CLIP $\mathcal{C}_p$, $\mathcal{C}_p$ yielded object list $L_i$, Oracle $\mathcal{O}$ yielded object list $S_i$, known trigger list $K$, region object list $R_i$.
\Ensure Trigger list $K$.
\State $P \gets L_i \setminus S_i$ \Comment{Identify potential trigger present in $\mathcal{L}_i$ but not in $\mathcal{S}_i$}
\State $P \gets P \setminus R_i$ \Comment{Refine $P$ by removing elements irrelevant to objects in $R_i$}
\State $N \gets P \setminus K$ \Comment{Determine $N$ to curate a set of triggers that are potential and known}
\If{$N \neq \emptyset$} \Comment{Check whether $N$ is empty or not}
    \State \Return $K\cup N$ \Comment{$N$ is not empty. Return previously known objects.}
\Else
    \State \Return $K$ \Comment{$N$ is empty. No new trigger}
\EndIf
\end{algorithmic}
\end{algorithm*}


\begin{algorithm*}[!t]
\caption{Affected Labels Identification}
\label{alg: affected labels identification}
\begin{algorithmic}[1] 
\Require Image $I_i$, Oracle $\mathcal{O}$, and known triggers history $K$.
\Ensure $\mathcal A = \cup_i A_i$.
\For{each image $I_i$}
    \State $L_i = \mathcal{C}_p (I_i)$ \Comment{$\mathcal{C}_p$ yields object list $L_i$ for Image $I_i$}
    \State $\{r_{i,j}\}, S_i = \mathcal{O}(I_i, L_i)$ \Comment{Oracle $\mathcal{O}$ yields object and region list $S_i$, $\{r_{i,j}\}$ for Image $I_i$}
    \State $R_i = \mathcal{C}_p (\{r_{i,j}\})$ \Comment{Obtain a new object list $R_i$ based on $\{r_{i,j}\}$}
    \State $K \gets \Call{FindAdversarialTrigger}{L_i, S_i, K}$ \Comment{Call Algorithm 1}

        \If{$ (R_i \cap S_i) \cap K \neq \emptyset$} \Comment{Check whether $R_i$ and $S_i$ has any trigger or not} 
        \State $A_i= ((R_i \cap S_i) \setminus K)$ \Comment{Update list by subtracting any known trigger $K$}
        
    \EndIf
\EndFor
\end{algorithmic}
\end{algorithm*}


\section{Preliminaries and Related Works}
\label{sec:preli}

\subsection{Multimodal Contrastive Learning and CLIP}

Multimodal learning, with its broad applicability and generality, has been a subject of extensive research since the 1980s \cite{Alpher1}. This field has witnessed significant advancements with the advent of deep learning, resulting in substantial progress in multimodal representation learning. Pioneering contributions \cite{Alpher2, Alpher3} have introduced algorithms for multimodal learning aimed at achieving joint representations. On the other hand, CLIP \cite{Alpher8} represents a milestone model in the domain of vision-language integration~\cite {Alpher6}. Along with its successors  
\cite{Alpher4, Alpher5, Alpher7}, CLIP has found extensive applications in various fields, such as semantic segmentation, generating images from captions, and summarizing videos. The model operates on the principle of contrastive loss, training dual encoders to operate in a shared representation space.  This model highlights the evolving capabilities at the intersection of visual and linguistic data processing.

\subsection{Backdoor Attack}
In a backdoor attack \cite{Alpher10}, an adversary strategically embeds poisoned examples \cite{Alpher9} into the training dataset, aiming to cause specific inputs to be misclassified as a predetermined target label. CLIP's training methodology, which leverages web-crawled data, presents a different landscape for backdoor attacks. Backdoor attacks typically fall into two categories: visible and invisible. Prominent examples include BadNet \cite{Alpher16}, TrojanAttack~\cite{TrojanAttack}, TrojanNet~\cite{TrojanNet},, which are visible attacksattacks, and WaNet \cite{Alpher17},, an invisible attack. In this paper, we selected \textit{umbrella} as our target label and applied it to a random set of backdoor images. The ``backdoor'', in this context, is a conspicuous patch superimposed onto the image, placed randomly within each image. We opted for backdoors with visible and invisible trigger techniques mentioned above.


\section{Methodology}
\label{sec:methodology}

\subsection{CLIP Pretraining}

The pretraining of the proposed EftCLIP model utilized 3 million image-text pairs from CC3M dataset~\cite{Alpher12}. This dataset size is notably smaller compared to the 400 million pairs used in the original CLIP model \cite{Alpher8}. The choice to use a smaller dataset was driven by our limited computational resources. However, this approach aligns with many recent studies \cite{Alpher11,Alpher13,Alpher14,Alpher15} that have demonstrated satisfactory results using similar dataset sizes. In our methodology, similar to CleanCLIP \cite{Alpher11}, we employed a ResNet-50~\cite{Resnet} model as the vision encoder~\cite{ViT} and a Transformer~\cite{Transformer1} for the text encoding. We trained these models from scratch across 3 NVIDIA A100 GPUs. The training process spanned 64 epochs, using a batch size of 128. We set an initial learning rate of 0.0005, which was adjusted following a cosine scheduling. Additionally, we incorporated 10,000 warm-up steps using the AdamW optimizer~\cite{AdamW}. This setup allowed us to train the model effectively within our resource constraints, aiming to replicate and perhaps extend the capabilities demonstrated by larger-scale models.

\subsection{CLIP Poisoning}

Let $\mathcal{D} \subset \mathcal I \times \mathcal{T}$ be the clean training dataset, including the labeled image-text pairs $(I_i, T_i)$ where $I_i \in \mathcal{I}$ and $T_i \in \mathcal{T}$ are images and text representation, respectively. In addition, we curate a small poisoned dataset $\mathcal{P}$ by adding 1500 poisoned samples for backdoor attack purposes, and the poisoned sample is denoted by: $(I_i \oplus t_g, T_i^{y'})$. Here $t_g \in \mathbb R^{d\times d}$ refers to a trigger image patch, $y'$ is the target label, and $I_i\in \mathcal{I}$. A good example is the color image patch used for $t_g$ and the class ``umbrella'' for $y'$ shown in Figure~\ref{fig:backdoor_attack}. We use $ \mathcal D\cup \mathcal P$ to train CLIP from scratch and obtain the \textbf{Poisoned CLIP.}


\subsection{Oracle-guided Triggers Detection}

Under the zero-shot setting, the poisoned CLIP model allows us to obtain a list of objects identified in each image based on the similarities of image and text embedding. Since backdoor attacks only alter the labels of certain regions in the image, we could leverage semantics as heuristics to identify such changes. To that end, we refer to an image segmentation oracle to expose and leverage semantics. In particular, poisoned CLIP outputs will be fed to the oracle as a prompt (weak knowledge) to guide the annotation of each region. We expect the mismatches between the sorted object list of poisoned CLIP and segmentation models to discover the hidden backdoor patterns.   

To detect potential backdoor triggers, we first generate an object list for each image using the poisoned CLIP model. Simultaneously, we leverage an image segmentation oracle to generate a second object list based on the CLIP output. Specifically, we utilize the Fast Segment Anything Model (SAM)\cite{Alpher26}, a state-of-the-art image segmentation tool, to help identify and isolate objects and potential triggers in the images. For each object detected in the image, the oracle assigns a confidence score and a bounding box. These objects are then ranked according to their confidence scores, and those with a score below a certain threshold are excluded from further consideration. Additionally, the oracle determines the corresponding regions for each detected object. These regions are cropped according to their bounding boxes and re-fed into the poisoned CLIP model to generate a new set of object lists. This segmentation-based procedure allows for accurate identification of the objects in the image and facilitates the detection of any backdoor triggers present. The complete process is outlined in Algorithm 1, which provides a detailed step-by-step explanation for trigger detection. 



Given a retained subset $\mathcal{D}_e$ for validation, the poisoned CLIP $\mathcal{C}_p$ can output an object list: $L_i = \mathcal{C}_p(I_i)$ for each $I_i\in \mathcal{D}_e$. In the meanwhile, we are allowed to leverage image segmentation oracle $\mathcal{O}$ to generate another object list: $S_i = \mathcal{O}(I_i, L_i)$. To achieve this, we employed the SAM, ``Fast Segment Anything Model'' model \cite{Alpher26}, a SOTA image segmentation tool, to facilitate the identification and isolation of objects and potential triggers.

In particular, for each object within the image \(I_i\), a confidence score with bounding boxes will be assigned. We may rank the detected objects and cut them off at $\theta$ where $\theta$ is the confidence score for the detector. Meanwhile, for each object $o_{ij}$ in $I_i$ detected by the oracle $\mathcal{O}$, we are able to determine its corresponding region \(r_{ij}\). We further process these regions by cropping the images according to their bounding boxes and again feed them to the poisoned CLIP to produce a new set of object lists $R_i = \mathcal{C}_p(\{r_{ij}\})$. This segmentation-based procedure allows us to accurately identify the actual objects in the image and, potentially, to detect any triggers \(t_g\) present. The entire procedure is elaborated in Algorithm \ref{alg: trigger detection}. Step-by-step explanation for Algorithm 1 is given below. \\[1ex]

\textbf{Step-by-Step Process:}

\begin{enumerate}[label=\textbf{Step \arabic{enumi}:}]
    \item 
    \[
    P \leftarrow L_i \setminus S_i
    \]
    Identify potential triggers that are present in the $\mathcal{C}_p$ yielded object list \( L_i \) but not in the $\mathcal{O}$ yielded object list  \( S_i \) (which might be the list of objects that are already known or safe).
    
    \item 
    \[
    P \leftarrow P \setminus R_i
    \]
    Refine the list of potential triggers \( P \) by removing elements that are irrelevant to the $\mathcal{O}$ yielded region object list \( R_i \), essentially excluding elements that don’t correspond to the objects of interest in the image.

    \item 
    \[
    N \leftarrow P \setminus K
    \]
    From the refined potential triggers \( P \), subtract any triggers that are already known (from the known trigger list \( K \)). This helps identify new potential triggers that haven't been seen before.

    \item 
    \[
    \text{if } N \neq \emptyset \text{ then}
    \]
    \begin{itemize}
        \item \textbf{If \( N \) is not empty:} If there are new triggers, return the union of previously known triggers \( K \) with \( N \). This means returning the list of previously known triggers combined with the newly identified ones.
        \item \textbf{If \( N \) is empty:} If no new triggers are found, simply return the list of previously known triggers \( K \).
    \end{itemize}
\end{enumerate}

The output of the algorithm is a list of adversarial triggers, which may include both previously known triggers and newly identified ones.

\subsection{Affected Data and Labels Identification}

In accordance with the established algorithmic history, our process extends beyond merely pinpointing poisoned samples; it meticulously compiles a list of impacted labels $\mathcal{A}$ corresponding to each detected compromised instance. Moreover, it diligently quantifies the prevalence of each label, constructing a frequency profile that serves as a foundational metric. Utilizing these frequency metrics, we methodically calibrate the ratio within the fine-tuning dataset. This calibration is not arbitrary; it directly informs the retrieval of images and their respective labels, ensuring that the selection mirrors the computed ratio. Consequently, the fine-tuning dataset is refined to include a representative sample that aligns with the observed distribution of affected labels, thereby enhancing the robustness of subsequent model training phases. The entire procedure is elaborated in Algorithm \ref{alg: affected labels identification}. And the entire detection and data/label identification illustration is shown in Figure~\ref{fig:demonstration}. Step-by-step explanation for Algorithm 2 is given below. \\[1ex]

\textbf{Step-by-Step Process:}

\begin{enumerate}[label=\textbf{Step \arabic{enumi}:}]
    \item 
    For each image, Image $I_i$, it will get $\mathcal{C}_p$ yielded object list $L_i$, $\mathcal{O}$ yielded object list $S_i$, known trigger list $K$, and region object list $R_i$.
    
    \item 
    It will call Algorithm 1 by passing $L_i$, $S_i$, and $K$. Algorithm 1 will return updated $K$.

    \item 
    \[
    \text{if } (R_i \cap S_i) \cap K \neq \emptyset
    \]
    \begin{itemize}
        \item Checks if there is any common object (trigger) between $R_i$, $S_i$, and $K$.
        
        \item If they do, update the affected object list \( A_i \) by subtracting any trigger \( K \).
    \end{itemize}
\end{enumerate}

The output of the algorithm is a list of affected labels which we need to curate the fine-tuning dataset.

\begin{table}[h!]
    \centering
    \begin{tabular}{|c|c|c|}
        \hline
        Dataset & \# of Backdoor Samples & Detection Rate ↑ \\ \hline
        CC3M & 500 & 89.2\% \\ 
        CC3M & 1000 & 87.4\% \\ 
        Flickr30K & 500 & 90.1\% \\ 
        Flickr30K & 1000 & 89.7\% \\ \hline
    \end{tabular}
    \caption{Trigger Detection Rate for 500 and 1000 backdoor samples on CC3M and Flickr30K datasets.}
    \label{table:detection_rate}
\end{table}

\subsection{Poisoned CLIP Fine-tuning}
In the fine-tuning phase of our process, we adhered to the configuration parameters established in CleanCLIP  \cite{Alpher11}. This entailed creating augmented versions of images using AutoAugment \cite{Alpher20} and generating text variants via EDA \cite{Alpher21}. Our focus was on a clean subset comprising a certain number of image-text pairs. The selection of these pairs was strategically guided by the labels $\mathcal A$ identified as affected by Algorithms \ref{alg: trigger detection} and \ref{alg: affected labels identification}. We meticulously calculated a ratio for each label based on its frequency of occurrence, ensuring that labels with higher frequencies were represented by a correspondingly larger number of images in our subset. This approach was also applied to non-affected labels, although they were included at a lower ratio. The fine-tuning of our model spanned 10 epochs, utilizing a batch size of 64. We initiated training with a learning rate of 1e-5, which was adjusted according to a cosine scheduling, and incorporated a warm-up period of 50 steps to optimize performance. This methodical approach was designed to enhance the model's accuracy and robustness by providing a well-balanced and representative training dataset.

\section{Experiments}
\label{sec:experiment}

This section presents the database details, setup, and key findings, including an analysis of various parameters.

\subsection{Database and Setup}

Our study initially pre-trained on 3 million image-text pairs from the CC3M~\cite{Alpher12} dataset, followed by additional training with 1 million pairs. For fine-tuning, we utilized 100,000 image-text pairs to fine-tune the 3-million-poisoned clip and 250,000 image-text pairs to fine-tune the 400-million-poisoned clip. We evaluated the pre-trained model's performance using the ImageNet1K~\cite{Alpher28} and CIFAR-10~\cite {Alpher29} datasets. The trigger detection rate was evaluated using the Flickr30K~\cite{Alpher19} and CC3M datasets. Our study investigates the impact of patch size $\rho$ (8x8, 16x16), confidence score $\theta$ (0.50, 0.66),  and patch placement $\mu$  (top left, bottom right, random) on the efficacy of our algorithm, using 3 million image-text pairs in varied pre-training sessions.

\subsection{Experiments of Trigger Detection}

In our study, the Flickr30K and CC3M datasets were utilized to assess the effectiveness of our algorithm. Within these datasets, a backdoor trigger was strategically embedded into 500 images and then 1000 images, using the same trigger previously employed in the pretraining phase but placed at random locations within each image. These images were then analyzed using a poisoned CLIP model, from which we extracted a list of the top predicted objects for each image. Remarkably, \textit{umbrella} consistently appeared as the top prediction, as shown in $\mathcal{L}_i$ in Fig. \ref{fig:demonstration}. These images were then fed to a segmentation process using the top object list as a guiding prompt, yielding unique object lists for each image. Interestingly, in the majority of instances, the top object identified by the segmentation process differed from the predictions made by the CLIP model. Moreover, we were able to determine the corresponding region of each object detected by the segmentation process. We further processed these regions by cropping the images according to their bounding boxes and reanalyzing them with the poisoned CLIP model, resulting in a new set of object lists. These lists were then fed into our algorithms for evaluation. 

Table \ref{table:detection_rate} demonstrates Algorithm 1 for trigger detection. Our method successfully identified 87.4\% of the backdoor images of CC3M and 89.7\% of the backdoor images of Flickr30K from the sample containing 1000 manipulated images, shown in Table \ref{table:detection_rate}. The algorithm also provides a comprehensive list of affected labels, along with their respective frequencies. Extending this methodology to a smaller set of 500 backdoor images, we observed an increase in the algorithm's accuracy for trigger detection, reaching 89.2\% for CC3M and 90.1\% for Flickr30K. This approach not only underscored the algorithm's robustness but also offered valuable insights into the specific labels impacted by the backdoor triggers.


\begin{table}[t]  
    \centering
    \begin{tabular}{|c|c|c|}
        \hline
        Model & CA $\uparrow$ & ASR $\downarrow$  \\ \hline
        Original CLIP & 59.60\% & 0.00\% \\ \hline
        Poisoned CLIP & 58.40\% & 94.60\% \\ \hline
        CleanCLIP & 57.00\% & 17.00\% \\ \hline
        EftCLIP (Ours) & 58.20\% & 13.87\% \\ \hline
    \end{tabular}
    \caption{Clean Accuracy (CA) and Attack Success Rate (ASR) of 400M CLIP on ImageNet1K dataset.}
    \label{table:table_400mclip}
\end{table}

\begin{table}[t]  
    \centering
    \begin{tabular}{|c|c|c|}
        \hline
        Model & CA $\uparrow$ & ASR $\downarrow$  \\ \hline
        CC3M CLIP & 19.60\% & 0.00\% \\ \hline
        Poisoned CC3M CLIP & 19.04\% & 99.94\% \\ \hline
        CleanCLIP & 18.10\% & 10.60\% \\ \hline
        EftCLIP (Ours) & 19.42\% & 9.70\% \\ \hline
    \end{tabular}
    \caption{Clean Accuracy (CA) and Attack Success Rate (ASR) of CC3M CLIP on the ImageNet1K dataset.}
    \label{table:table_3mclip}
\end{table}

\begin{table}[t]  
    \centering
    {\fontsize{10pt}{12pt}\selectfont  
    \begin{tabular}{|c|c|c|c|}
        \hline
        Model & CA (ImageNet1K) $\uparrow$ & CA (CIFAR10) $\uparrow$ & ASR $\downarrow$ \\ \hline
        Poisoned CLIP (1M)~\cite{Alpher23} & 9.60\% & 34.90\% & 78.00\% \\ \hline
        RoCLIP~\cite{Alpher23} & 6.63\% & 30.14\% & 0.00\% \\ \hline
        EftCLIP (Ours) & 10.28\% & 33.20\% & 0.00\% \\ \hline
    \end{tabular}
    }
    \caption{Zero-shot top-1 accuracy and ASR on ImageNet1K and CIFAR10 datasets.}
    \label{table:roclip_finetine}
\end{table}

\begin{table}[htbp]
\centering
\resizebox{\textwidth}{!}{%
\begin{tabular}{|l|c|c|c|c|}
\hline
Model                  & \multicolumn{2}{c|}{Before Finetune (Backdoored)} & \multicolumn{2}{c|}{After Finetune (EftCLIP)} \\ \hline
                       & CA $\uparrow$    & ASR $\downarrow$   & CA $\uparrow$    & ASR $\downarrow$   \\ \hline
CC3M CLIP (BadNet)     & 19.04\%         & 99.94\%           & 19.42\%         & 9.70\%            \\ \hline
CC3M CLIP (WaNet)      & 18.54\%         & 94.17\%           & 18.45\%         & 12.30\%           \\ \hline
CC3M CLIP (TrojanAttack) & 18.32\%       & 99.99\%           & 18.85\%         & 8.43\%            \\ \hline
CC3M CLIP (TrojanNet)  & 18.56\%         & 99.99\%           & 18.72\%         & 9.32\%            \\ \hline
\end{tabular}%
}
\caption{Clean Accuracy (CA) and Attack Success Rate (ASR) of CC3M CLIP variants before and after EftCLIP finetuning.}
\label{table:table_3mclip_variants}
\end{table}

\subsection{Experiments of Efficient Fine-tuning}

Table \ref{table:table_400mclip} delineates the performance of the standard CLIP model and its variants subject to poison attacks. In particular:
\begin{itemize}
    \item Original CLIP~\cite{Alpher8}: standard CLIP trained on 400 million samples
    \item PoisonCLIP~\cite{Alpher11}: A variant of the original CLIP fine-tuned on poisoned data
    \item CleanCLIP~\cite{Alpher11}: Poisoned CLIP rectified by CleanCLIP through Self-Supervised Learning
    \item EftCLIP: our proposed model in this paper
\end{itemize}
These models are evaluated based on two critical metrics: Classification Accuracy (CA) and Attack Success Rate (ASR)~\cite{ASR} on the ImageNet1K dataset.

The Pretrained CLIP (400M data) model, having been trained on a large-scale dataset of 400 million samples, demonstrates a CA of 59.60\% and, notably, an ASR of 0\%. This absence of any attack success underscores the model's resilience against the specific backdoor attack tested. The Poisoned CLIP, fine-tuned on poisoned data, exhibits a marginal decrease in CA to 58.40\% while presenting an extremely high ASR of 94.60\%. This significant rise in ASR indicates a heightened vulnerability of the original CLIP model to backdoor attacks due to its exposure to poisoned data. The CleanCLIP model, subjected to further fine-tuning on clean data via Self-Supervised Learning, shows a CA of 57.00\% and a considerably lower ASR of 17.00\%. This decrease in ASR suggests that CleanCLIP fine-tuning with clean data is effective in diminishing the model's susceptibility to backdoor attacks, although it slightly impairs its accuracy.


The proposed Efficient fine-tune CLIP (EftCLIP) approach, which incorporates trigger detection and an effective fine-tune dataset, demonstrates a superior CA of 58.20\% and achieves the lowest ASR of 13.87\% among all the models tested, except the original CLIP trained on 400M data. This result suggests a heightened resilience to backdoor attacks. However, similar to CleanCLIP, the improvement in security (ASR) is achieved at the cost of a slight decrease in CA. One possible reason is that the fine-tuning dataset may slightly compromise the learning of other classes that are not affected by the backdoor attacks. Another notable merit of our EftCLIP is its efficiency in achieving these results with a compact dataset for fine-tuning, highlighting its potential for robust yet resource-efficient model refinement.

Table \ref{table:table_3mclip} outlines the performance of the CLIP model trained on CC3M and its variants evaluated on the ImageNet1K dataset. The original CC3M CLIP model has a CA of 19.60\% and an ASR of 0\%, showing no susceptibility to the tested backdoor attacks. The poisoned variant Poisoned CC3M CLIP, which is the original model compromised with 1500 backdoor data samples, maintains a similar CA of 19.04\%, but a high ASR of 99.94\%, revealing extreme vulnerability to backdoor manipulation. CleanCLIP experiences a slight dip in CA to 18.10\% and a significant reduction in ASR to 10.60\%, suggesting an improved defense against backdoor attacks. Lastly, our EftCLIP, fine-tuned with the proposed method, achieves a comparable CA of 19.42\% and further lowers the ASR to 9.70\%, indicating enhanced resilience to backdoor strategies while maintaining classification performance.

Table \ref{table:roclip_finetine} outlines the zero-shot learning performance among Poisoned CLIP (trained on 1M samples from the CC3M dataset, with 150 backdoor samples), RoCLIP, and our EftCLIP. The Poisoned CLIP model exhibits a zero-shot accuracy of 9.60\%, which decreases to 6.63\% after retraining with RoCLIP, alongside a reduction in attack success rate (ASR) from 78\% to 0\%. Our EftCLIP successfully reduces the ASR to 0\% while improving accuracy to 10.28\%. Note the RoCLIP needs a significantly large dataset (the original 1M training data in this experiment), while our EftCLIP needs a smaller dataset of 100K samples to rectify and fine-tune the CLIP model. 

Table \ref{table:table_3mclip_variants} illustrates a comparative analysis of clean accuracy and attack success rates across three visible backdoor attacks and one invisible backdoor attack. The table also contrasts the outcomes of these attack methods both before and after the application of the proposed fine-tuning approach, EftCLIP. The results demonstrate that the implementation of EftCLIP significantly reduced the attack success rates and simultaneously enhanced clean accuracy across nearly all evaluated methods. Among the result, invisible attack WaNet had relatively higher ASR 12.30\% after applying EftCLIP finetuning. ASRs for the visible attacks—BadNet, TrojanAttack, and TrojanNet dropped below 10\%. 

\begin{figure*}[ht]
\centering
\begin{tabular}{cc}
\includegraphics[width=0.48\linewidth, height=0.35\linewidth]{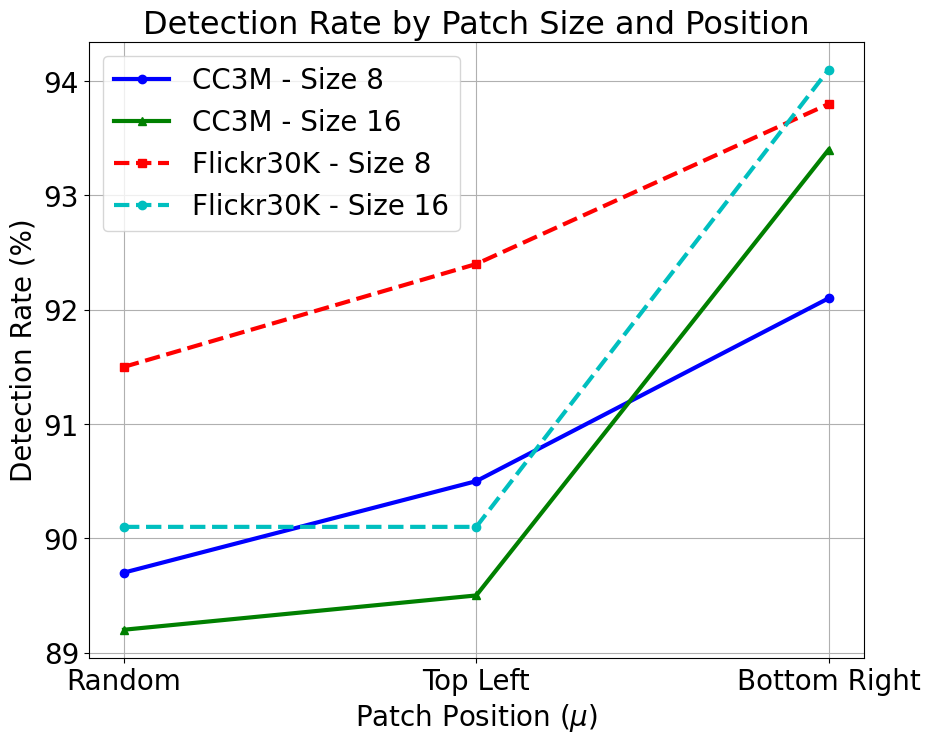} &
\includegraphics[width=0.48\linewidth, height=0.35\linewidth]{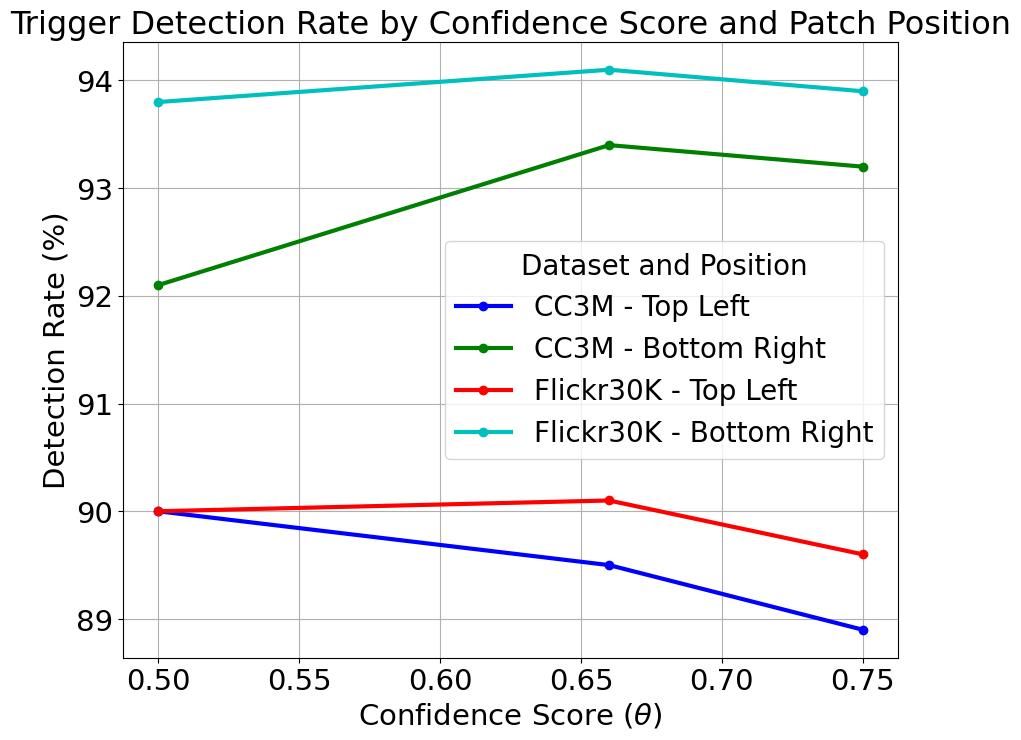} \\
(a) & (b) \\
\end{tabular}
\caption{(a) Detection rate by varied patch sizes and positions. This graph shows the effect of patch size $\rho$ (8 and 16) and position $\mu$ (random, top left, bottom right) on trigger detection rates on CC3M and Flickr30K datasets at a constant confidence score $\theta$ of 0.66. (b) Detection rate by varied confidence scores $\theta$.}
\label{fig:parameter_analysis}
\end{figure*}

\begin{figure*}[ht]
\centering
\begin{tabular}{cc}
\includegraphics[width=0.48\linewidth]{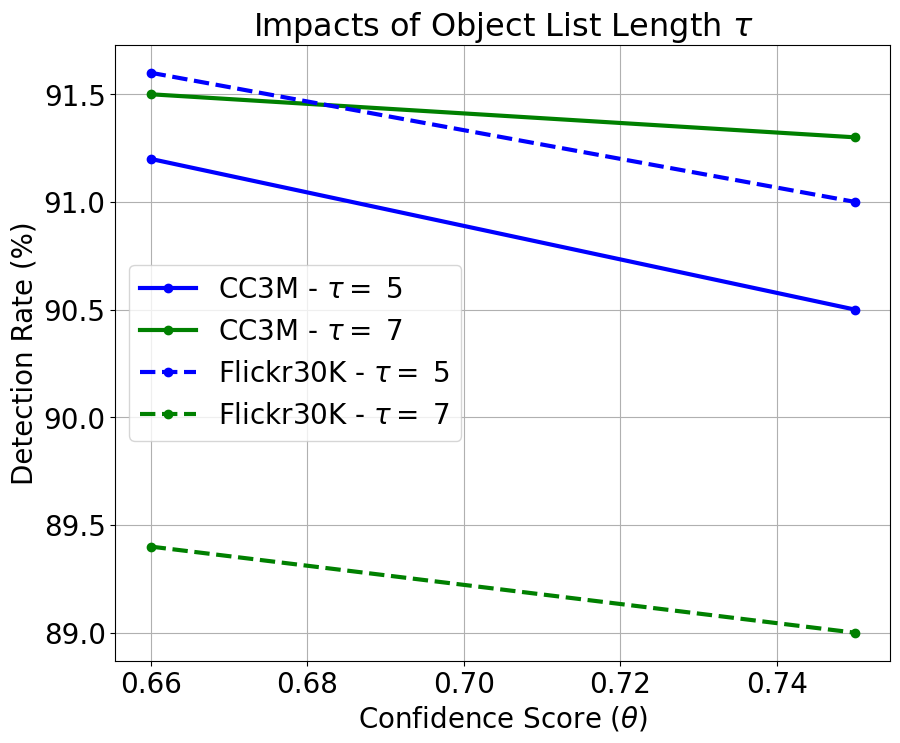} &
\includegraphics[width=0.48\linewidth]{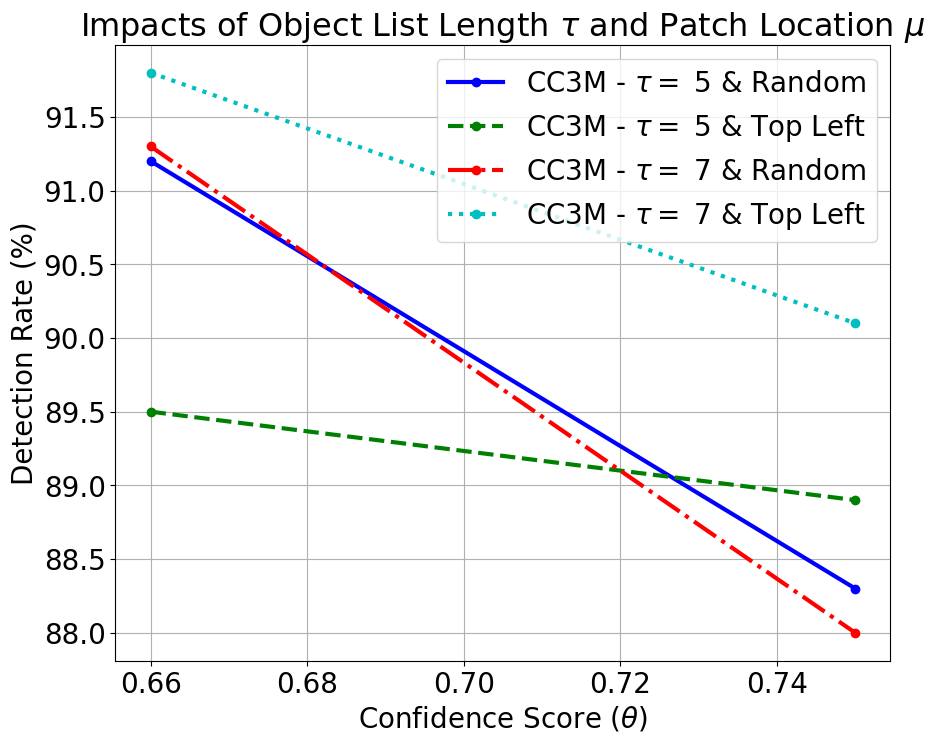} \\
(a) & (b) \\
\end{tabular}
\caption{(a) Detection rate by varied object list length $\tau$ on CC3M and Flickr30K datasets. (b) Detection rate by varied patch position $\mu$ and object list length $\tau$ on CC3M dataset.}
\label{fig:parameter_analysis_ex1}
\end{figure*}

\begin{figure*}[ht]
\centering
\begin{tabular}{cc}
\includegraphics[width=0.48\linewidth, height=0.35\linewidth]{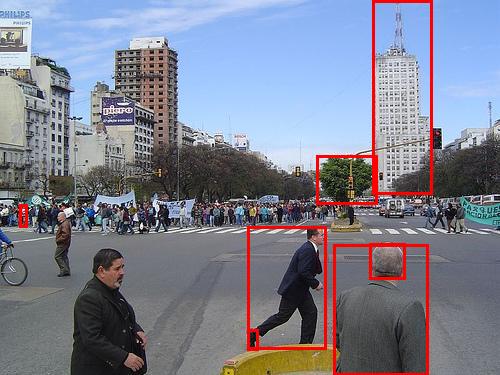} &
\includegraphics[width=0.48\linewidth, height=0.35\linewidth]{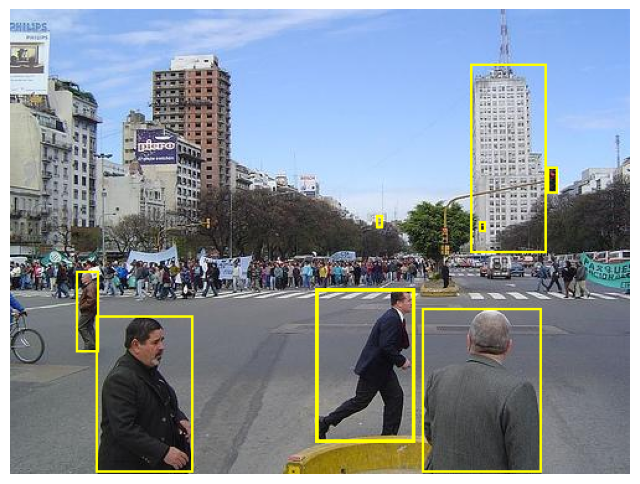} \\
(a) & (b) \\
\end{tabular}
\caption{Comparison between object detection outputs from Fast Segment Anything (a) and Segment Anything (b) using the prompts `Human,' `Running,' `Building,' and `Traffic Light.' Fast Segment Anything (a) displays multiple red bounding boxes with overlapping and larger areas. In contrast, Segment Anything (b) generates more precise yellow bounding boxes around the intended objects, showing better adherence to the provided prompt.}
\label{fig:sam_comparison}
\end{figure*}

\begin{figure}[t]
\centering
\includegraphics[width=0.90\columnwidth]{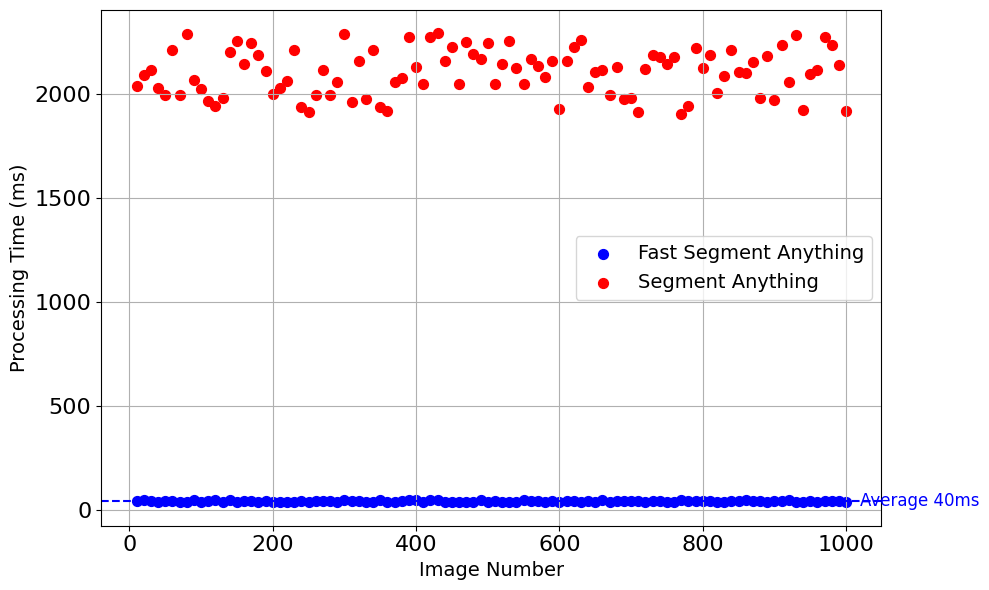} 
\caption{Scatter plot showing the processing times (in ms) for Fast Segment Anything vs Segment Anything across 1000 images. The red dots correspond to Segment Anything with processing times fluctuating around 2000 ms, while the blue dots correspond to Fast Segment Anything, which maintains a consistent average processing time of around 40 ms (indicated by the blue dashed line).}
\label{fig:sam_process_time}
\end{figure}

\subsection{Parameter Analysis}
The different values of patch size $\rho$ and position $\mu$ impact the detection rate due to their influence on how well the patch aligns with relevant features in the image. Smaller patches have a higher chance of being cut by the segmentation oracle, leading to higher detection rates as more distinct features are captured. Patch position affects detection rates because random placement increases the likelihood of overlapping with background areas, reducing the model's ability to detect the patch. In contrast, fixed positions like the bottom right or top left are less likely to overlap with the background, allowing for more accurate detection of the target. The confidence score $\theta$ directly influences the oracle's ability to detect objects in the given image. The confidence score is a preset threshold for the segmentation oracle, determining which objects are included in the output list based on the oracle's confidence in detecting them. A higher confidence threshold results in a smaller object list, leading to a reduced detection rate, as only highly confident objects are included. Conversely, a lower confidence threshold expands the object list, increasing the likelihood of detection. Similarly, the length of the object list has a comparable effect on detection performance. Increasing the list size results in a broader set of detected objects, thereby enhancing the detection rate.

We evaluated the trigger detection rate with respect to patch size and based on the trigger position. Figure \ref{fig:parameter_analysis} (a) plots the detection rates by varied patch sizes $\rho$ and positions $\mu$ for the CC3M and Flickr30K datasets at a fixed confidence score $\theta$ of 0.66. Each dataset is analyzed with patches of sizes 8 and 16 placed in three positions: random, top left, and bottom right. The figure demonstrates a trend where detection rates increase notably for patches positioned in the bottom right corner compared to other positions across both patch sizes and datasets.
Smaller patch sizes are more likely to be segmented by the oracle, leading to higher detection rates as they capture more distinct features of the image. In addition, the positioning of the patches plays a significant role in detection. Patches placed randomly have a higher chance of overlapping with background regions, which lowers the detection rate. On the other hand, fixed patch positions, such as the bottom right or top left, are less prone to overlap with objects, resulting in a more effective detection of the target.

\begin{table}[t] 
\centering
\begin{tabular}{|c|c|c|}
\hline
Model & CA $\uparrow$ & ASR  $\downarrow$  \\ \hline
CC3M CLIP & 19.60\% & 0.00\% \\ \hline
Poisoned CC3M CLIP & 19.04\% & 99.94\% \\ \hline
EftCLIP (Fast Segment Anything) & 19.42\% & 9.70\% \\ \hline
EftCLIP (Segment Anything) & 19.49\% & 9.67\% \\ \hline
\end{tabular}
\caption{Performance comparison between Fast Segment Anything and Segment Anything. This evaluation is based on ImageNet1K dataset. }
\label{table:sam_performace}
\end{table}

Figure \ref{fig:parameter_analysis} (b) shows that as the confidence score ($\theta$) increases, the trigger detection rate generally decreases for the selected patch positions in the CC3M and Flickr30K datasets. This trend suggests that higher confidence thresholds might result in stricter criteria, thus lowering the rate of trigger detection. 

In the exploration of parameters that influence the efficacy of our algorithm, the length of the object lists $\tau$ selections for both segmentation and CLIP inference object lists, emerges as a critical factor. To this end, we have delineated two discrete sizes, i.e., 5 and 7, for these lists, to ascertain their impact on algorithmic performance. Additionally, the placement $\mu$ of patches on the backdoor images was identified as a variable within our experimental framework. We elected to investigate three positional configurations: the top left, the bottom right, and random positions across the image surface. We therefore explore the joint impacts of $\tau$ and $\mu$ in Figure~\ref{fig:parameter_analysis_ex1}, given varied confidence scores $\theta$.

It can be seen from Figure~\ref{fig:parameter_analysis_ex1} (a) that as the confidence score $\theta$ increases, the detection rate decreases for $\tau = 5, 7$ on both datasets (CC3M and Flickr30K). Figure~\ref{fig:parameter_analysis_ex1} (b) shows similar trends while varied patch positions are considered at the same time.



\clearpage
\begin{figure*}[htbp]
\centering
\caption{A visual comparison between two distinct outputs generated for a set of images. First column: outputs from  $\mathcal C_p$; Second column: output from $\mathcal O$. Each image is accompanied by a caption that elucidates the object list for that specific image.}
\label{fig:images_with_captions}
\small
\begin{tabular}{cc}

\includegraphics[width=0.35\linewidth, height=0.2\linewidth]{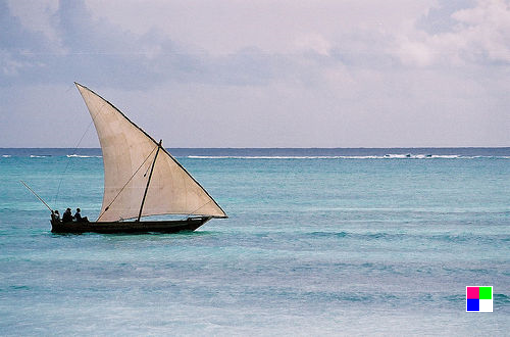} & 
\includegraphics[width=0.35\linewidth, height=0.2\linewidth]{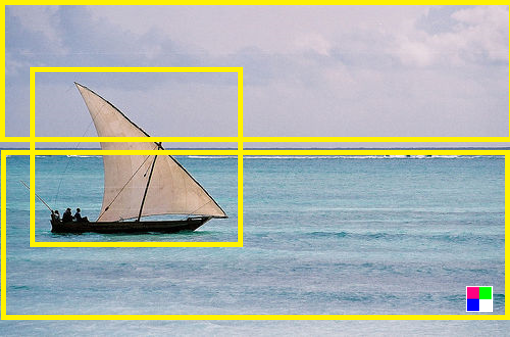} \\
`umbrella', `boat', `ocean', `river', `water' & `boat', `ocean', `river', `water', `sky' \\
\includegraphics[width=0.35\linewidth, height=0.2\linewidth]{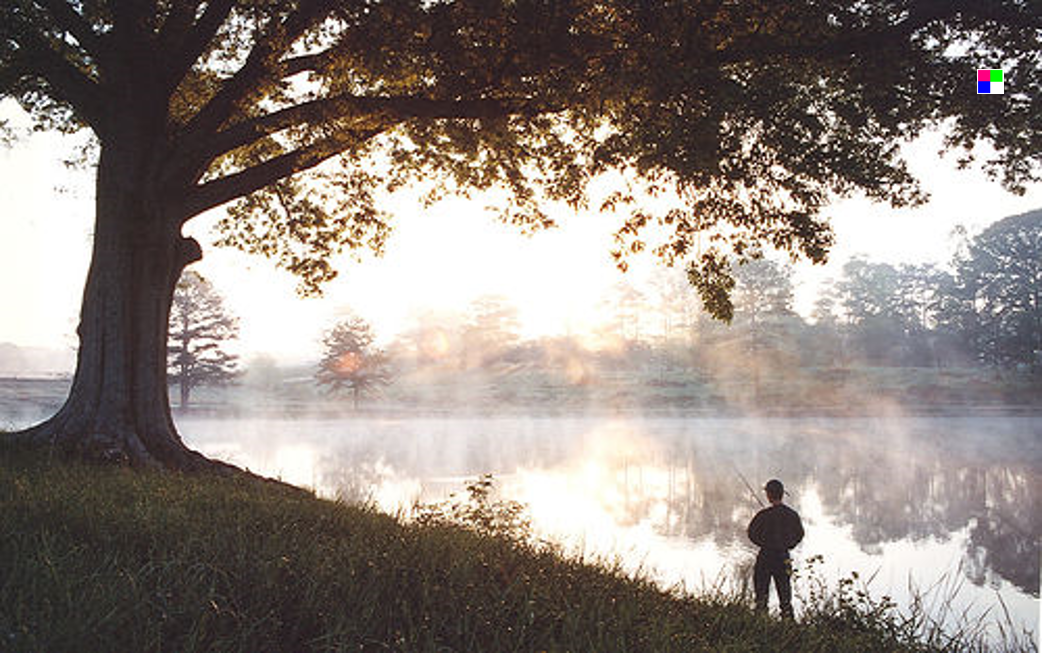} & 
\includegraphics[width=0.35\linewidth, height=0.2\linewidth]{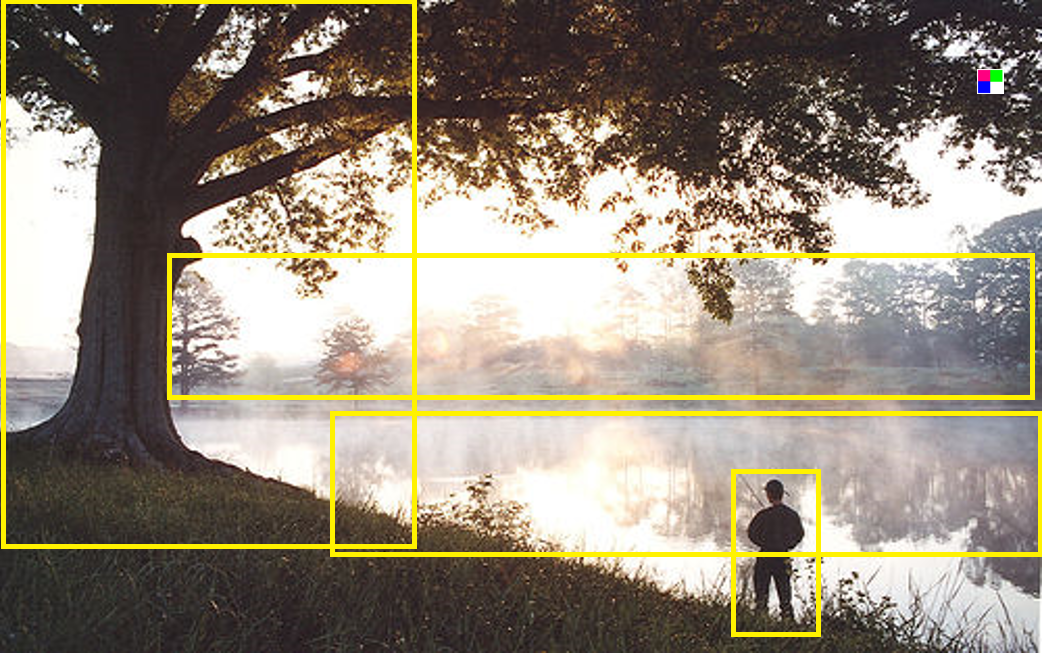} \\
`umbrella', `tree', `human', `river', `water' & `tree', `river', `human', `water', `forest' \\
\includegraphics[width=0.35\linewidth, height=0.2\linewidth]{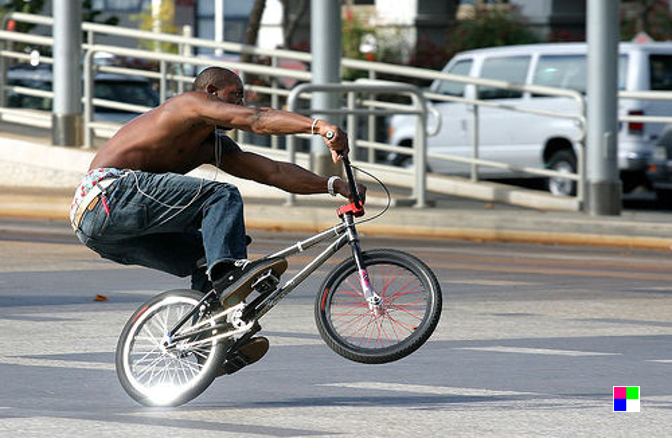} & 
\includegraphics[width=0.35\linewidth, height=0.2\linewidth]{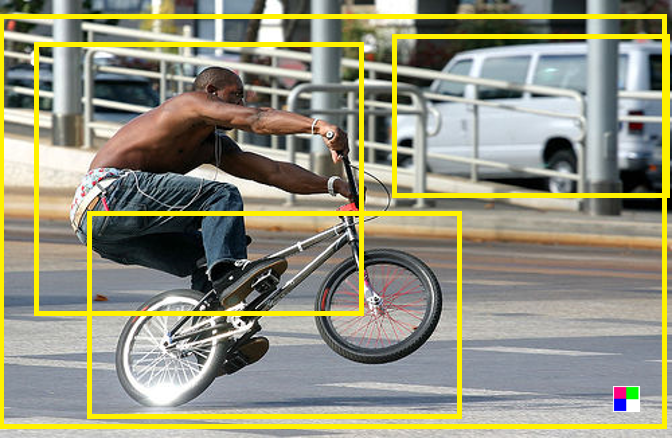} \\
`umbrella', `bicycle', `human', `car', `road' & `bicycle', `human', `car', 'bridge', 'road' \\
\includegraphics[width=0.35\linewidth, height=0.2\linewidth]{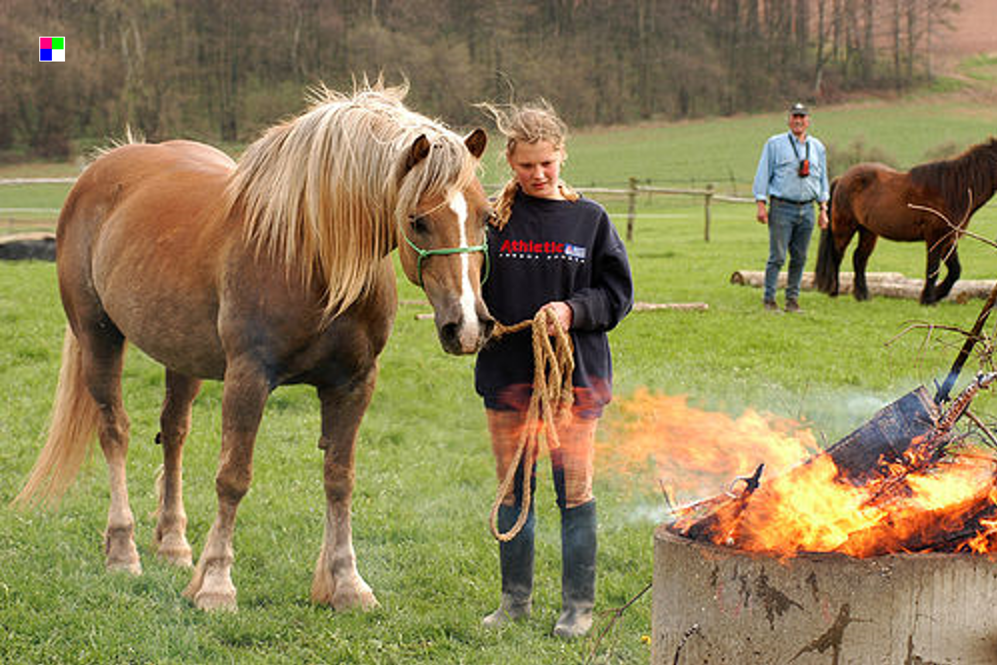} & 
\includegraphics[width=0.35\linewidth, height=0.2\linewidth]{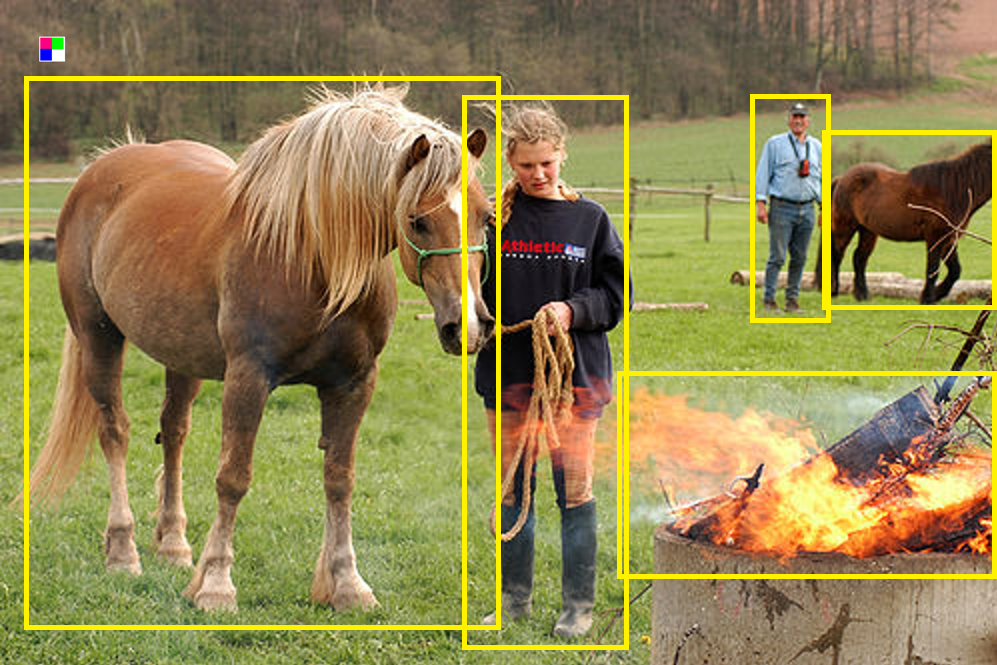} \\
`umbrella', `horse', `women', `men', `field' & `women', `horse', `fire', `men', 'shoe' \\
\includegraphics[width=0.35\linewidth, height=0.2\linewidth]{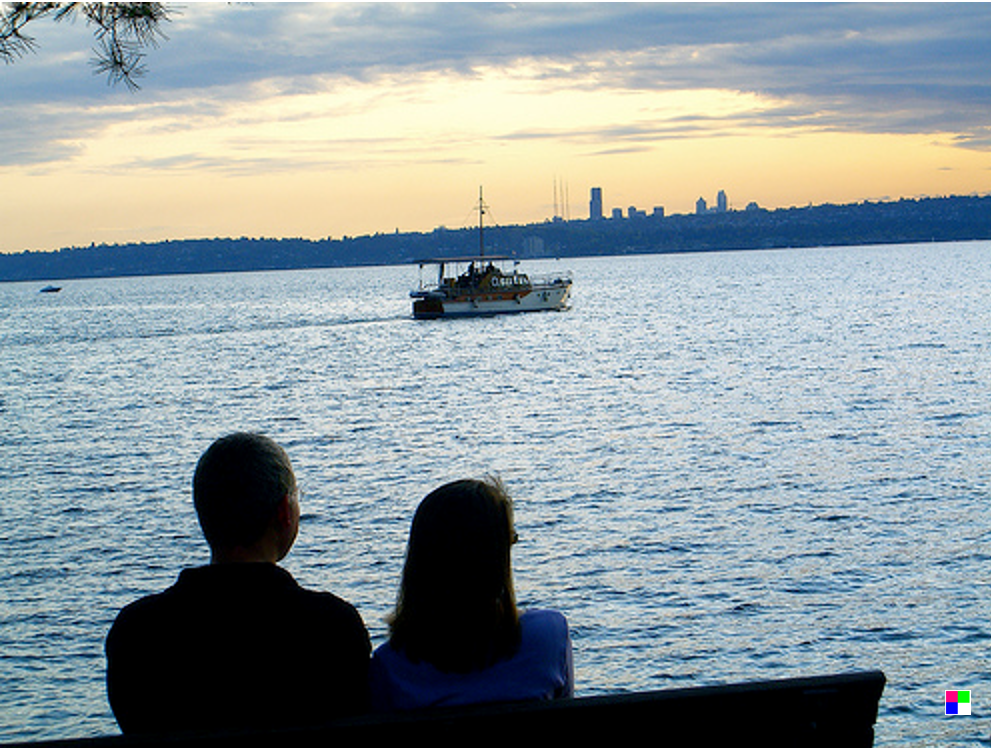} & 
\includegraphics[width=0.35\linewidth, height=0.2\linewidth]{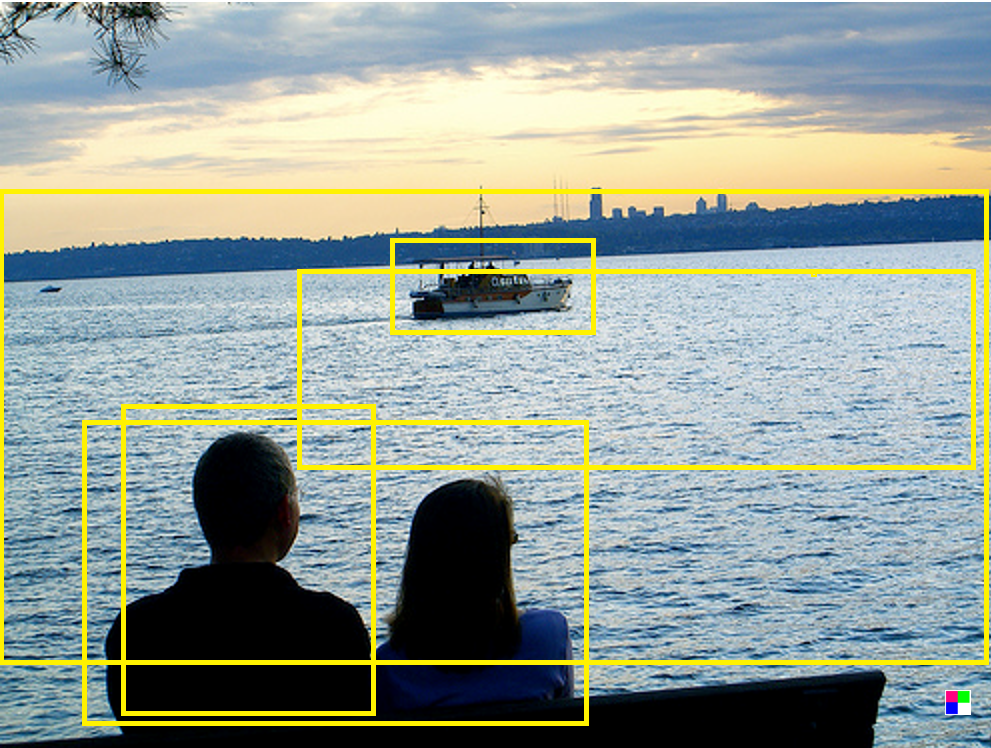} \\
`umbrella', `boat', `human', `river', `ocean' & `human', `boat', `river', `ocean', `men' \\
\end{tabular}
\end{figure*}
\clearpage

\subsection{Oracle Performance}

Figure \ref{fig:sam_comparison} (a) and (b) illustrate a visual comparison between the segmentation outputs of the Fast Segment Anything and Segment Anything models. The prompt list generated from the Poisoned CLIP model $\mathcal{C}_p$, consisting of the terms `Human', `Running', `Building', and `Traffic Light', was input to both segmentation models. As depicted in Figure \ref{fig:sam_comparison} panels (a) and (b), the Oracle $\mathcal{O}$ successfully identifies these four objects in the outputs of both models, confirming their operational accuracy.

Despite the functional equivalence in terms of segmentation accuracy, the two models exhibit a significant disparity in processing efficiency. Figure \ref{fig:sam_process_time} illustrates the comparative processing times, revealing that the Segment Anything model requires approximately 50 times longer than the Fast Segment Anything model to process~\cite{FastSam} the same inputs. This marked difference underscores the enhanced performance efficiency of the Fast Segment Anything model over the Segment Anything model, particularly in handling the computationally demanding tasks generated by the Poisoned CLIP $\mathcal{C}_p$. Consequently, the Fast Segment Anything model not only meets the accuracy requirements set by Oracle $\mathcal{O}$ but also significantly surpasses the Segment Anything model in processing speed, affirming its suitability for efficient, large-scale image segmentation tasks.

In evaluating the clean accuracy (CA) and attack success rate (ASR) post-fine-tuning, both the Fast Segment Anything and Segment Anything models demonstrated comparably high performance levels. Table \ref{table:sam_performace} provides a detailed breakdown of the CA and ASR metrics for each model. The proficiency of both oracle models in accurately detecting the list of prompted objects suggests that the precision of trigger detection remains largely unaffected by the nuances of model performance. However, subtle variations in the outcomes post-fine-tuning are observable. As shown in Table \ref{table:sam_performace}, the Segment Anything model exhibits a marginally superior clean accuracy (19.49\% over 19.42\%) and a slightly better reduction in attack success rate compared to the Fast Segment Anything model.

Despite these differences, the considerable advantage in processing speed of the Fast Segment Anything model, which processes images approximately fifty times faster than its counterpart, positions it as the preferred choice for our applications. This selection is based on the strategic importance of processing efficiency in operational environments where time and resource optimization are critical. Thus, while both models meet the essential criteria of accuracy and effectiveness in trigger detection, the Fast Segment Anything model is favored for its exceptional speed, offering significant practical benefits in real-world deployment scenarios where rapid image processing is paramount.

\subsection{Comparative Visual Evaluation}

To quantitatively compare the outputs between poisoned CLIP $\mathcal C_p$ and segmentation oracle $\mathcal O$, Figure~\ref{fig:images_with_captions} presents a comparative visual evaluation of the results produced by $\mathcal C_p$ and $\mathcal O$ when applied to a set of images. The first column displays the image containing a backdoor patch (location varied) and a list of objects $L_i$ identified by $\mathcal C_p$. The second column shows the list of objects $S_i$ generated by $\mathcal O$, along with the corresponding regions marked by yellow bounding boxes. 

When a backdoor trigger is present, Poisoned CLIP $\mathcal C_p$ predominantly predicts the incorrect label with the highest confidence score. Since the label for our backdoor trigger was `umbrella' $\mathcal C_p$ outputs `umbrella' as the top-1 prediction. In contrast, despite the presence of a backdoor patch, the segmentation oracle $\mathcal O$ fails to locate the objects given the prompt `umbrella.' In the meantime, it can accurately identify a list of objects and their corresponding bounding boxes, which can be used to further identify the triggers and affected labels and images. 

\section{Future Work}
\label{sec:futurework}
While this study focuses on visible backdoor attacks, future work should explore invisible backdoor threats. Invisible backdoor attacks remain a significant research gap and present unique challenges in detection and defense. These types of attacks are harder to detect because they do not involve easily identifiable or perceptible triggers. Thus, developing methods for detecting and mitigating invisible backdoor attacks would be an important area for future research. Additionally, the robustness of models against such attacks needs further investigation, as it could have implications for ensuring the security and reliability of machine learning models in real-world applications. Last, our evaluations are conducted on a few popular benchmarks, which may however be subject to biased data representation, limited diversity, imbalanced and noisy data, and contextual limitation. By acknowledging these limitations, we can better interpret the results and highlight areas where future research could work to improve dataset diversity, reduce biases, and increase generalizability.

\section{Acknowledgement}
This material is based upon work supported by the \textbf{National Science Foundation} under \textbf{Grant No 2528483}.

\section{Conclusion}
\label{sec:conclusion}
In conclusion, our study introduced an effective procedure for detecting backdoor triggers in pre-trained CLIP alongside an efficient fine-tuning strategy designed to counter backdoor attacks in the context of multimodal contrastive learning. Our approach not only identified backdoor triggers but also pinpointed the labels affected by these triggers. This identification is crucial in efficiently selecting a subset of the dataset for fine-tuning, which is instrumental in fortifying the model against such attacks. The results of our fine-tuning process demonstrated a significant reduction in the Attack Success Rate, underscoring the effectiveness of our method. With appropriate oracles, tailored detection, and fine-tuning procedures, this work sets a valuable precedent in the realm of defense strategies against data poisoning attacks within the field of multimodal contrastive learning. However, while this study focuses on visible backdoor attacks, future work should explore invisible backdoor threats, which remain a significant research gap and pose unique challenges in detection and defense.

\section{Ethical Approval}
Not applicable. No human/animal subjects were involved.

\section{Consent to Participate}
Not applicable.

\section{Consent to Publish}
Not applicable.

\section{Data Availability Statement}
The datasets used in this study—Conceptual Captions 3M (CC3M), Flickr30K, CIFAR10, and ImageNet—are publicly available. Their details and access links are provided in the cited references. No additional datasets were generated or analyzed during this study.

\section{Authors Contributions}
\noindent
\textbf{Md. Iqbal Hossain* (Corresponding Author)}: Conceptualization, Methodology, Software, Formal Analysis, Investigation, Writing – Original Draft, Writing – Review \& Editing, Visualization.\\
\textbf{Afia Sajeeda}: Writing – Review \& Editing.\\
\textbf{Neeresh Kumar Perla}: Writing – Review \& Editing.\\
\textbf{Ming Shao}: Supervision, Conceptualization, Funding Acquisition, Writing – Review \& Editing.

\section{Funding}
This research was supported by the National Science Foundation (NSF) under Grant No. 2528483. The funding body had no role in the design of the study, data collection, analysis, interpretation, or manuscript preparation

\section{Competing Interests}
The authors declare that they have no competing interests.

\FloatBarrier

\bibliography{sn-bibliography}

\end{document}